\documentclass[letterpaper, 10 pt, conference]{ieeeconf}  

\IEEEoverridecommandlockouts                              

\usepackage{graphics} 
\usepackage{epsfig} 
\usepackage{amsmath} 
\usepackage{amssymb}  
\usepackage{dsfont}
\usepackage{hyperref}
\usepackage{multirow}
\usepackage{booktabs,caption}
\usepackage[flushleft]{threeparttable}
\usepackage{float}

\title{\LARGE \bf
SqueezeSeg: Convolutional Neural Nets with Recurrent CRF for Real-Time Road-Object Segmentation from 3D LiDAR Point Cloud
}

\author{Bichen Wu, Alvin Wan, Xiangyu Yue and Kurt Keutzer\\
UC Berkeley\\
\{bichen, alvinwan, xyyue, keutzer\}@berkeley.edu
}

\begin{document}

\maketitle
\thispagestyle{empty}
\pagestyle{empty}

\begin{abstract}
In this paper, we address semantic segmentation of road-objects from 3D LiDAR point clouds. In particular, we wish to detect and categorize instances of interest, such as cars, pedestrians and cyclists. We formulate this problem as a point-wise classification problem, and propose an end-to-end pipeline called SqueezeSeg based on convolutional neural networks (CNN): the CNN takes a transformed LiDAR point cloud as input and directly outputs a point-wise label map, which is then refined by a conditional random field (CRF) implemented as a recurrent layer. Instance-level labels are then obtained by conventional clustering algorithms. Our CNN model is trained on LiDAR point clouds from the KITTI~\cite{KITTI} dataset, and our point-wise segmentation labels are derived from 3D bounding boxes from KITTI. To obtain extra training data, we built a LiDAR simulator into \textit{Grand Theft Auto V (GTA-V)}, a popular video game, to synthesize large amounts of realistic training data. Our experiments show that SqueezeSeg achieves high accuracy with astonishingly fast and stable runtime ($8.7\pm 0.5$ ms per frame), highly desirable for autonomous driving applications. Furthermore, additionally training on synthesized data boosts validation accuracy on real-world data. Our source code and synthesized data will be open-sourced.
\end{abstract}

\section{INTRODUCTION}

Autonomous driving systems rely on accurate, real-time and robust perception of the environment. An autonomous vehicle needs to accurately categorize and locate ``road-objects'', which we define to be driving-related objects such as cars, pedestrians, cyclists, and other obstacles. Different autonomous driving solutions may have different combinations of sensors, but the 3D LiDAR scanner is one of the most prevalent components. LiDAR scanners directly produce distance measurements of the environment, which are then used by vehicle controllers and planners. Moreover, LiDAR scanners are robust under almost all lighting conditions, whether it be day or night, with or without glare and shadows. As a result, LiDAR based perception tasks have attracted significant research attention. 

\begin{figure}[t]
    \centering
    \includegraphics[width=3.4in,trim={0cm 0cm 0cm 0cm},clip]{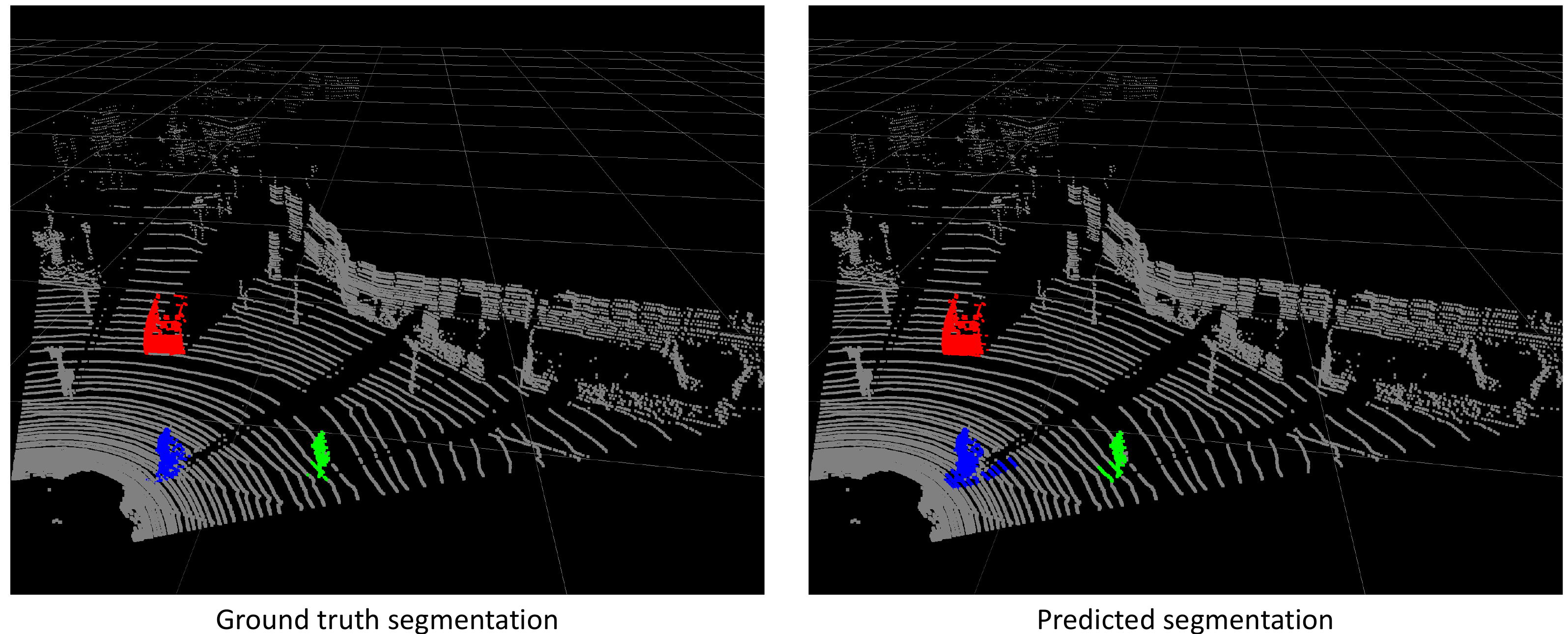}
    \caption{An example of SqueezeSeg segmentation results. Our predicted result is on the right and the ground truth is on the left. Cars are annotated in red, pedestrians in green and cyclists in blue.}
    \label{fig:LiDAR}
\end{figure}

In this work, we focus on road-object segmentation using (Velodyne style) 3D LiDAR point clouds. Given point cloud output from a LiDAR scanner, the task aims to isolate objects of interest and predict their categories, as shown in Fig. \ref{fig:LiDAR}. Previous approaches comprise or use parts of the following stages: Remove the ground, cluster the remaining points into instances, extract (hand-crafted) features from each cluster, and classify each cluster based on its features. This paradigm, despite its popularity~\cite{LiDARSegICRA2012,himmelsbach2008lidar,wang2012could,zermas2017fast}, has several disadvantages: a) Ground segmentation in the above pipeline usually relies on hand-crafted features or decision rules -- some approaches rely on a scalar threshold~\cite{thrun2006stanley} and others require more complicated features such as surface normals~\cite{LocalConvexitySeg} or invariant descriptors~\cite{wang2012could}, 
all of which may fail to generalize and the latter of which require significant preprocessing. b) Multi-stage pipelines see aggregate effects of compounded errors, and classification or clustering algorithms in the pipeline above are unable to leverage context, most importantly the immediate surroundings of an object. c) Many approaches for ground removal rely on iterative algorithms such as RANSAC (random sample consensus)~\cite{zermas2017fast}, GP-INSAC (Gaussian Process Incremental Sample Consensus)~\cite{LiDARSegICRA2012}, or agglomerative clustering~\cite{LiDARSegICRA2012}. 
The runtime and accuracy of these algorithmic components depend on the quality of random initializations and, therefore, can be unstable. This instability is not acceptable for many embedded applications such as autonomous driving. We take an alternative approach: use deep learning to extract features, develop a single-stage pipeline and thus sidestep iterative algorithms.

In this paper, we propose an end-to-end pipeline based on convolutional neural networks (CNN) and conditional random field (CRF). CNNs and CRFs have been successfully applied to segmentation tasks on 2D images ~\cite{FCN,CRF, DeepLab, CRFasRNN}. To apply CNNs to 3D LiDAR point clouds, we designed a CNN that accepts transformed LiDAR point clouds and outputs a point-wise map of labels, which is further refined by a CRF model. Instance-level labels are then obtained by applying conventional clustering algorithms (such as DBSCAN) on points within a category. To feed 3D point clouds to a 2D CNN, we adopt a spherical projection to transform sparse, irregularly distributed 3D point clouds to dense, 2D grid representations. The proposed CNN model draws inspiration from SqueezeNet~\cite{SqueezeNet} and is carefully designed to reduce parameter size and computational complexity, with an aim to  reduce memory requirements and achieve real-time inference speed for our target embedded applications. The CRF model is reformulated as a recurrent neural network (RNN) module as~\cite{CRFasRNN} and can be trained end-to-end together with the CNN model. Our model is trained on LiDAR point clouds from the KITTI dataset~\cite{KITTI} and point-wise segmentation labels are converted from 3D bounding boxes in KITTI. To obtain even more training data, we leveraged \textit{Grand Theft Auto V (GTA-V)} as a simulator to retrieve LiDAR point clouds and point-wise labels. 

\begin{figure*}[h]
    \centering
    \includegraphics[width=7in]{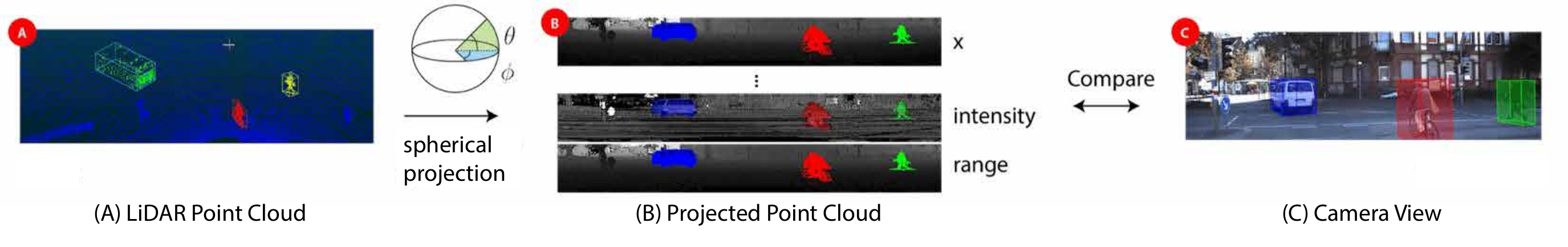}
    \caption{LiDAR Projections. Note that each channel reflects structural information in the camera-view image.}
    \label{fig:ProjectedData}
\end{figure*}

Experiments show that SqueezeSeg achieves high accuracy and is extremely fast and stable, making it suitable for autonomous driving applications. We additionally find that supplanting our dataset with artificial, noise-injected simulation data further boosts validation accuracy on real-world data.

\section{Related Work}
\subsection{Semantic segmentation for 3D LiDAR point clouds}
Previous work saw a wide range of granularity in LiDAR segmentation, handling anything from specific components to the whole pipeline.
\cite{LocalConvexitySeg} proposed mesh based ground and object segmentation relying on local surface convexity conditions.  
\cite{LiDARSegICRA2012} summarized several approaches based on iterative algorithms such as RANSAC (random sample consensus) and GP-INSAC (gaussian process incremental sample consensus) for ground removal. 
Recent work also focused on algorithmic efficiency. \cite{zermas2017fast} proposed efficient algorithms for ground segmentation and clustering while \cite{shin2017real} bypassed ground segmentation to directly extract foreground objects. 
\cite{wang2012could} expanded its focus to the whole pipeline, including segmentation, clustering and classification. It proposed to directly classify point patches into background and foreground objects of different categories then use EMST-RANSAC ~\cite{zermas2017fast} to further cluster instances. 

\subsection{CNN for 3D point clouds}
CNN approaches consider LiDAR point clouds in either two or three dimensions. Work with two-dimensional data considers raw images with projections of LiDAR point clouds top-down~\cite{CNNRoadSeg} or from a number of other views~\cite{MultiView}. Other work considers three-dimensional data itself, discretizing the space into voxels and engineering features such as disparity, mean, and saturation~\cite{LiDARFusion}. Regardless of data preparation, deep learning methods consider end-to-end models that leverage 2D convolutional~\cite{LiDARDet} or 3D convolutional~\cite{3DCNN} neural networks.

\subsection{Semantic Segmentation for Images}
Both CNNs and CRFs have been applied to semantic segmentation tasks for images. \cite{FCN} proposed transforming CNN models, trained for classification, to fully convolutional networks to predict pixel-wise labels. \cite{CRF} proposed a CRF formulation for image segmentation and solved it approximately with the mean-field iteration algorithm. CNNs and CRFs are combined in~\cite{DeepLab}, where the CNN is used to produce an initial probability map and the CRF is used to refine and restore details. In \cite{CRFasRNN}, mean-field iteration is re-formulated as a recurrent neural network (RNN) module.  

\subsection{Data Collection through Simulation}
Obtaining annotations, especially point-wise or pixel-wise annotations for computer vision tasks is usually very difficult. As a consequence, synthetic datasets have seen growing interest. In the autonomous driving community, the video game Grand Theft Auto has been used to retrieve data for object detection and segmentation~\cite{eccv_playing_for_data,drive_in_matrix}. 

\section{Method description}
\subsection{Point Cloud Transformation}
\label{sec:point_cloud}
Conventional CNN models operate on images, which can be represented by 3-dimentional tensors of size $H\times W \times 3$. The first two dimensions encode spatial position, where $H$ and $W$ are the image height and width, respectively. The last dimension encodes features, most commonly RGB values. However, a 3D LiDAR point cloud is usually represented as a set of cartesian coordinates, $(x, y, z)$. Extra features can also be included, such as intensity or RGB values. Unlike the distribution of image pixels, the distribution of LiDAR point clouds is usually sparse and irregular. Therefore, naively discretizing a 3D space into voxels results in excessively many empty voxels. Processing such sparse data is inefficient, wasting computation.

To obtain a more compact representation, we project the LiDAR point cloud onto a sphere for a dense, grid-based representation as
\begin{gather}
\label{eqn:projection}
\begin{split}
\theta = \arcsin{\frac{z}{\sqrt{x^2+y^2+z^2}}}, \ 
\tilde{\theta} = \lfloor\theta / \triangle \theta\rfloor, \\
\phi = \arcsin{\frac{y}{\sqrt{x^2+y^2}}}, \ 
\tilde{\phi} = \lfloor\phi / \triangle \phi\rfloor.
\end{split}
\end{gather}
$\phi$ and $\theta$ are \textit{azimuth} and \textit{zenith} angles, as shown in Fig. \ref{fig:ProjectedData} (A).  $\triangle \theta$ and $\triangle \phi$ are resolutions for discretization and $(\tilde{\theta}, \tilde{\phi})$ denotes the position of a point on a 2D spherical grid. Applying equation (\ref{eqn:projection}) to each point in the cloud, we can obtain a 3D tensor of size $H\times W\times C$. In this paper, we consider data collected from a Velodyne HDL-64E LiDAR with 64 vertical channels, so $H=64$. Limited by data annotations from the KITTI dataset, we only consider the front view area of $90^{\circ}$ and divide it into 512 grids so $W=512$.  $C$ is the number of features for each point. In our experiments, we used 5 features for each point: 3 cartesian coordinates $(x, y, z)$, an intensity measurement and range $r=\sqrt{x^2+y^2+z^2}$. An example of a projected point cloud can be found at Fig. \ref{fig:ProjectedData} (B). As can be seen, such representation is dense and regularly distributed, resembling an ordinary image Fig. \ref{fig:ProjectedData} (C). This featurization allows us to avoid hand-crafted features, bettering the odds that our representation generalizes.

\subsection{Network structure}
\begin{figure*}[h]
    \centering
    \includegraphics[width=6in]{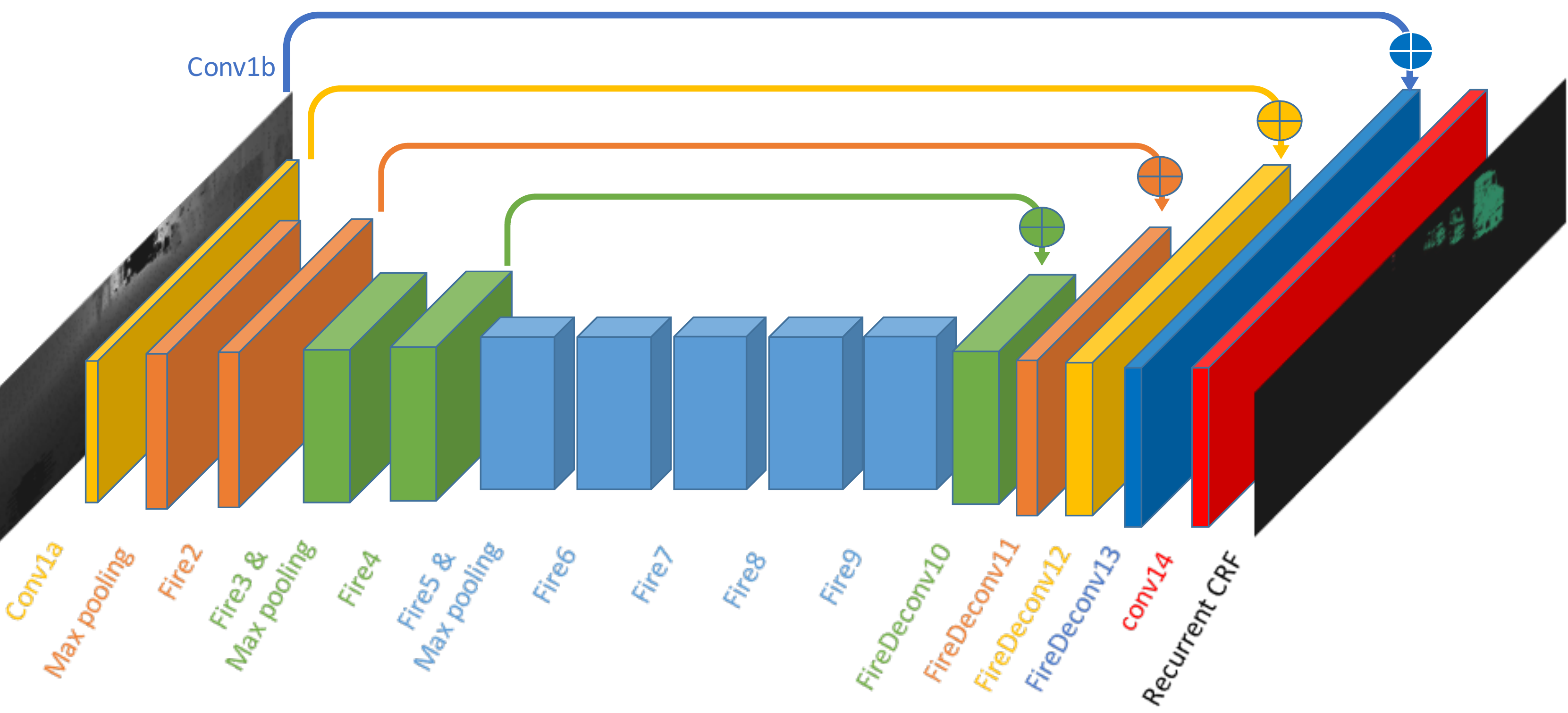}
    \caption{Network structure of SqueezeSeg.}
    \label{fig:NNstructure}
\end{figure*}

Our convolutional neural network structure is shown in Fig. \ref{fig:NNstructure}. SqueezeSeg is derived from SqueezeNet~\cite{SqueezeNet}, a light-weight CNN that achieved AlexNet~\cite{alexnet} level accuracy with 50X fewer parameters. 

The input to \textit{SqueezeSeg} is a $64\times 512 \times 5$ tensor as described in the previous section. We ported layers (\textit{conv1a} to \textit{fire9}) from SqueezeNet for feature extraction. SqueezeNet used \textit{max-pooling} to down-sample intermediate feature maps in both width and height dimensions, but since our input tensor's height is much smaller than its width, we only down-sample the width. The output of \textit{fire9} is a down-sampled feature map that encodes the semantics of the point cloud. 

To obtain full resolution label predictions for each point, we used deconvolution  modules (more precisely, ``transposed convolutions'') to up-sample feature maps in the width dimension. We used skip-connections to add up-sampled feature maps to lower-level feature maps of the same size, as shown in Fig. \ref{fig:NNstructure}. The output probability map is generated by a convolutional layer (\textit{conv14}) with \textit{softmax} activation. The probability map is further refined by a recurrent CRF layer, which will be discussed in the next section. 

\begin{figure}[h]
    \centering
    \includegraphics[width=2in]{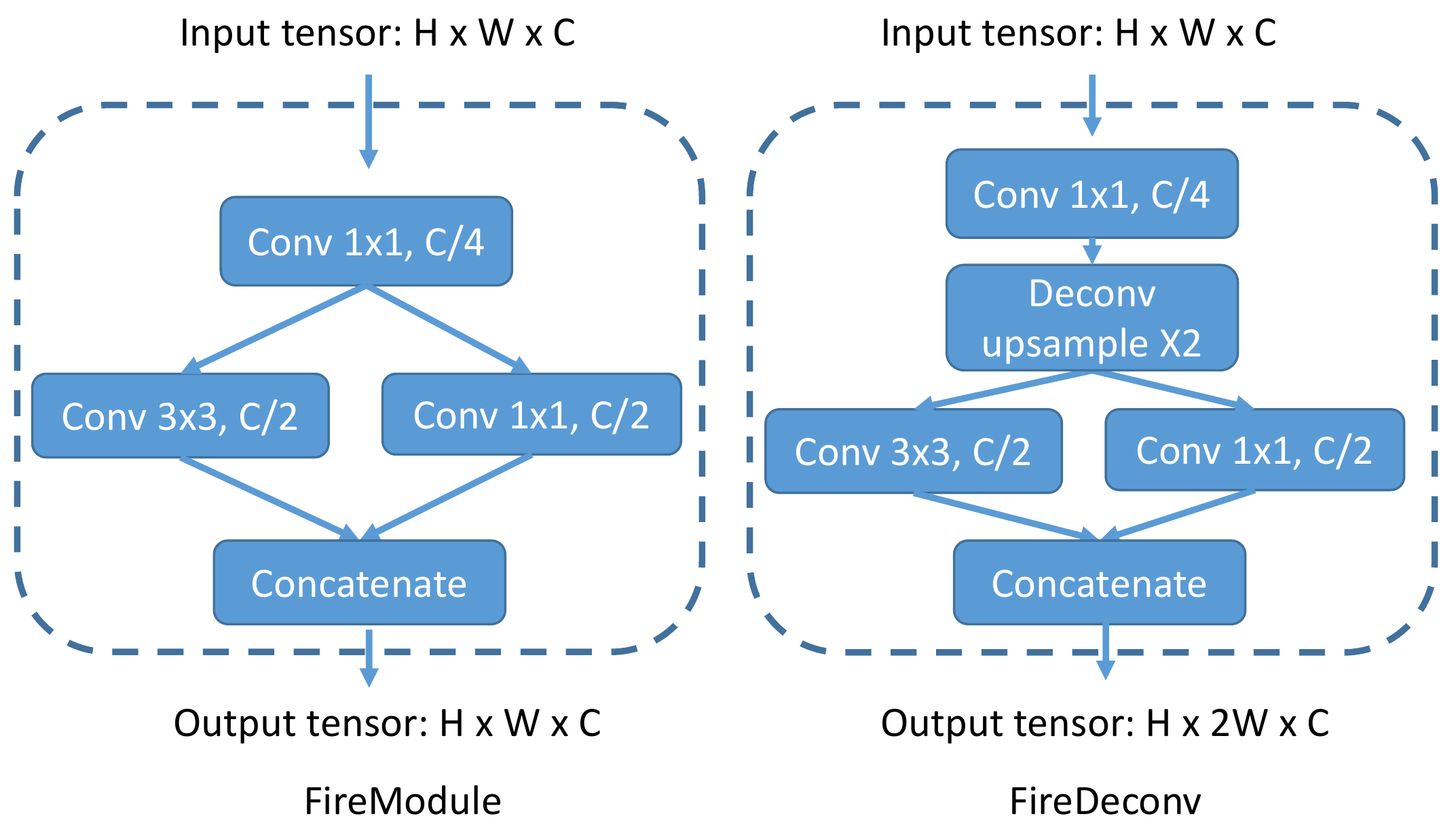}
    \caption{Structure of a \textit{FireModule} (left) and a \textit{fireDeconv} (right).}
    \label{fig:FireDeconv}
\end{figure}

In order to reduce the number of model parameters and computation, we replaced convolution and deconvolution layers with \textit{fireModule}s~\cite{SqueezeNet} and \textit{fireDeconv}s. The architecture of both modules are shown in Fig. \ref{fig:FireDeconv}. In a \textit{fireModule}, the input tensor of size $H\times W \times C$ is first fed into a 1x1 convolution to reduce the channel size to $C/4$. Next, a 3x3 convolution is used to fuse spatial information. Together with a parallel 1x1 convolution, they recover the channel size of $C$. The input 1x1 convolution is called the \textit{squeeze} layer and the parallel 1x1 and 3x3 convolution together is called the \textit{expand} layer. Given matching input and output size, a 3x3 convolutional layer requires $9C^2$ parameters and $9HWC^2$ computations, while the \textit{fireModule} only requires $\frac{3}{2}C^2$ parameters and $\frac{3}{2}HWC^2$ computations. In a \textit{fireDeconv} module, the deconvolution layer used to up-sample the feature map is placed between \textit{squeeze} and \textit{expand} layers. To up-sample the width dimension by 2, a regular 1x4 deconvolution layer must contain $4C^2$ parameters and $4HWC^2$ computations. With the \textit{fireDeconv} however, we only need $\frac{7}{4}C^2$ parameters and $\frac{7}{4}HWC^2$ computations.

\subsection{Conditional Random Field}
With image segmentation, label maps predicted by CNN models tend to have blurry boundaries. This is due to loss of low-level details in down-sampling operations such as max-pooling. Similar phenomena are also observed with SqueezeSeg.

Accurate point-wise label prediction requires understanding not only the high-level semantics of the object and scene but also low-level details. The latter are crucial for the consistency of label assignments. For example, if two points in the cloud are next to each other and have similar intensity measurements, it is likely that they belong to the same object and thus have the same label. Following~\cite{DeepLab}, we used a conditional random field (CRF) to refine the label map generated by the CNN. For a given point cloud and a label prediction $\mathbf{c}$ where $c_i$ denotes the predicted label of the $i$-th point, a CRF model employs the energy function
\begin{equation}
E(\mathbf{c}) = \sum_i u_i(c_i) + \sum_{i, j} b_{i, j}(c_i, c_j).
\label{eqn:CRF}
\end{equation}
The unary potential term $u_i(c_i) = -\log{P(c_i)}$ considers the predicted probability $P(c_i)$ from the CNN classifier. The binary potential terms define the ``penalty'' for assigning different labels to a pair of similar points and is defined as $b_{i,j}(c_i, c_j) = \mu(c_i, c_j) \sum_{m=1}^M w_m k^m(\mathbf{f_i}, \mathbf{f_j})$ where $\mu(c_i, c_j) = 1$ if $c_i \neq c_j$ and 0 otherwise, $k^m$ is the $m$-th Gaussian kernel that depends on features $\mathbf{f}$ of point $i$ and $j$ and $w_m$ is the corresponding coefficient. In our work, we used two Gaussian kernels
\begin{equation}
\begin{gathered}
w_1\exp (-\frac{\|\mathbf{p_i} - \mathbf{p_j}\|^2}{2\sigma_\alpha^2} 
       -\frac{\|\mathbf{x_i} - \mathbf{x_j}\|^2}{2\sigma_\beta^2}) \\
+ w_2\exp (-\frac{\|\mathbf{p_i} - \mathbf{p_j}\|^2}{2\sigma_\gamma^2}).
\end{gathered}
\label{eqn:kernels}
\end{equation}
The first term depends on both angular position $\mathbf{p} (\tilde{\theta}, \tilde{\phi})$ and cartesian coordinates $\mathbf{x}(x, y, z)$ of two points. The second term only depends on angular positions. $\sigma_\alpha$, $\sigma_\beta$ and $\sigma_\gamma$ are three hyper parameters chosen empirically. Extra features such as intensity and RGB values can also be included. 

\begin{figure}[h]
    \centering
    \includegraphics[width=3.5in]{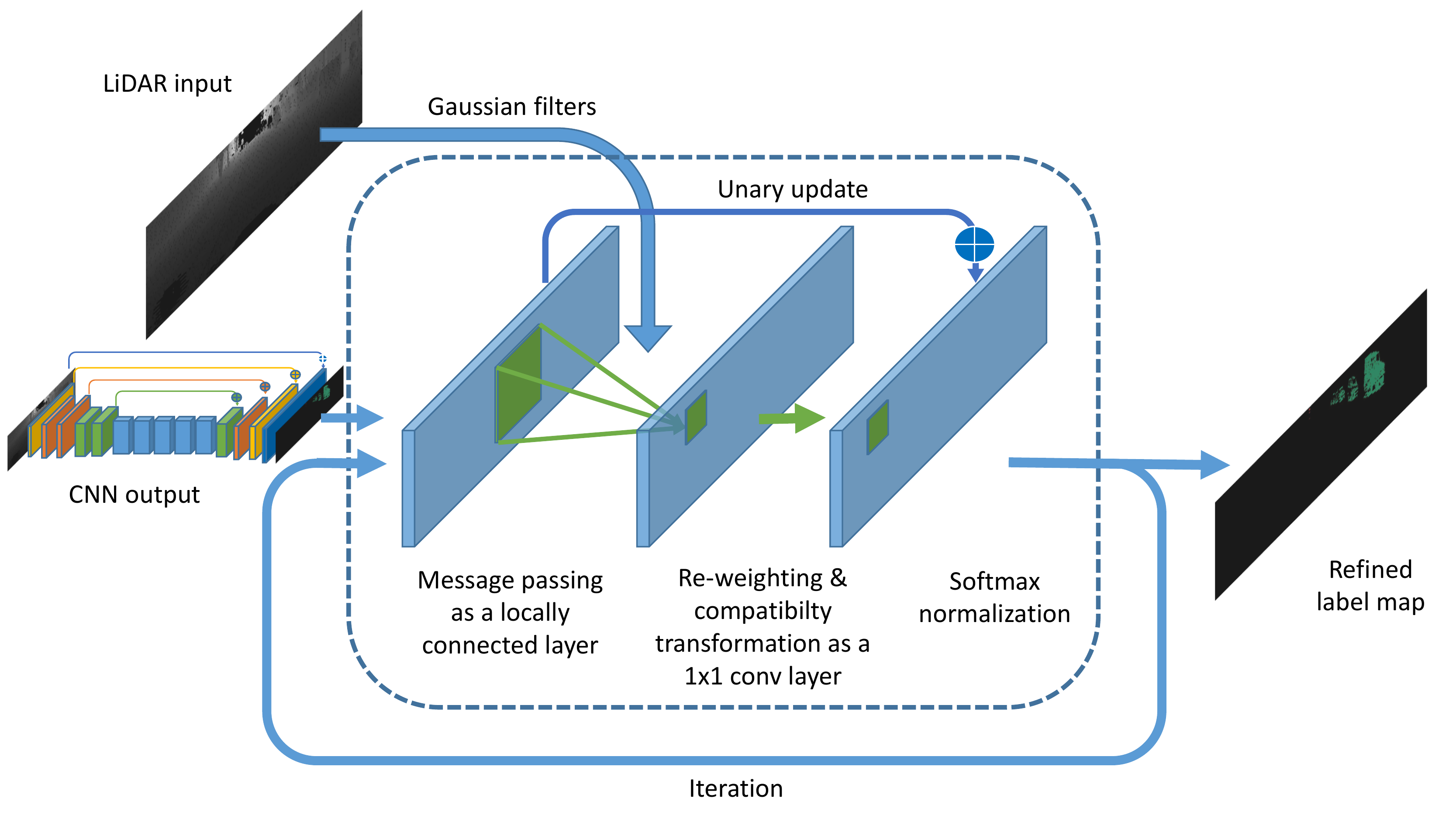}
    \caption{Conditional Random Field (CRF) as an RNN layer.}
    \label{fig:CRF}
\end{figure}

Minimizing the above CRF energy function yields a refined label assignment. Exact minimization of equation (\ref{eqn:CRF}) is intractable, but~\cite{CRF} proposed a mean-field iteration algorithm to solve it approximately and efficiently. ~\cite{CRFasRNN} reformulated the mean-field iteration as a recurrent neural network (RNN). We refer readers to \cite{CRF} and \cite{CRFasRNN} for a detailed derivation of the mean-field iteration algorithm and its formulation as an RNN. Here, we provide just a brief description of our implementation of the mean-field iteration as an RNN module as shown in Fig.~\ref{fig:CRF}. The output of the CNN model is fed into the CRF module as the initial probability map. Next, we compute Gaussian kernels based on the input feature as equation (\ref{eqn:kernels}). The value of above Gaussian kernels drop very fast as the distance (in the 3D cartesian space and the 2D angular space) between two points increases. Therefore, for each point, we limit the kernel size to a small region of $3\times 5$ on the input tensor. Next, we filter the initial probability map using above Gaussian kernels. This step is also called message passing in \cite{CRFasRNN} since it essentially aggregates probabilities of neighboring points. This step can be implemented as a locally connected layer with above Guassian kernels as parameters. Next, we re-weight the aggregated probability and use a ``compatibilty transformation'' to decide how much it changes each point's distribution. This step can be implemented as a 1x1 convolution whose parameters are learned during training. Next, we update the initial probability by adding it to the output of the 1x1 convolution and use \textit{softmax} to normalize it. The output of the module is a refined probability map, which can be further refined by applying this procedure iteratively. In our experiment, we used 3 iterations to achieve an accurate label map. This recurrent CRF module together with the CNN model can be trained together end-to-end. With a single stage pipeline, we sidestep the thread of propagated errors present in multi-stage workflows and leverage contextual information accordingly.

\subsection{Data collection}
Our initial data is from the KITTI raw dataset, which provides images, LiDAR scans and 3D bounding boxes organized in sequences. Point-wise annotations are converted from 3D bounding boxes. All points within an object's 3D bounding box are considered part of the target object. We then assign the corresponding label to each point. An example of such a conversion can be found in Fig. \ref{fig:ProjectedData} (A, B).
Using this approach, we collected 10,848 images with point-wise labels. 

In order to obtain more training samples (both point clouds and point-wise labels), we built a LiDAR simulator in GTA-V. The framework of the simulator is based on DeepGTAV\footnote{\url{https://github.com/ai-tor/DeepGTAV}}, which uses Script Hook V\footnote{\url{http://www.dev-c.com/gtav/scripthookv/}} as a plugin. 

We mounted a virtual LiDAR scanner atop an in-game car, which is then set to drive autonomously. The system collects both LiDAR point clouds and the game screen. In our setup, the virtual LiDAR and game camera are placed at the same position, which offers two advantages: First, we can easily run sanity checks on the collected data, since the points and images need to be consistent. Second, the points and images can be exploited for other research projects, e.g. sensor fusion, etc. 

We use ray casting to simulate each laser ray that LiDAR emits. The direction of each laser ray is based on several parameters of the LiDAR setup: vertical field of view (FOV), vertical resolution, pitch angle, and the index of the ray in the point cloud scan. Through a series of APIs, the following data associated with each ray can be obtained: a) the coordinates of the first point the ray hits, b) the class of the object hit, c) the instance ID of the object hit (which is useful for instance-wise segmentation, etc.), d) the center and bounding box of the object hit. 

\begin{figure}[h]
    \centering
    \includegraphics[width=3.4in]{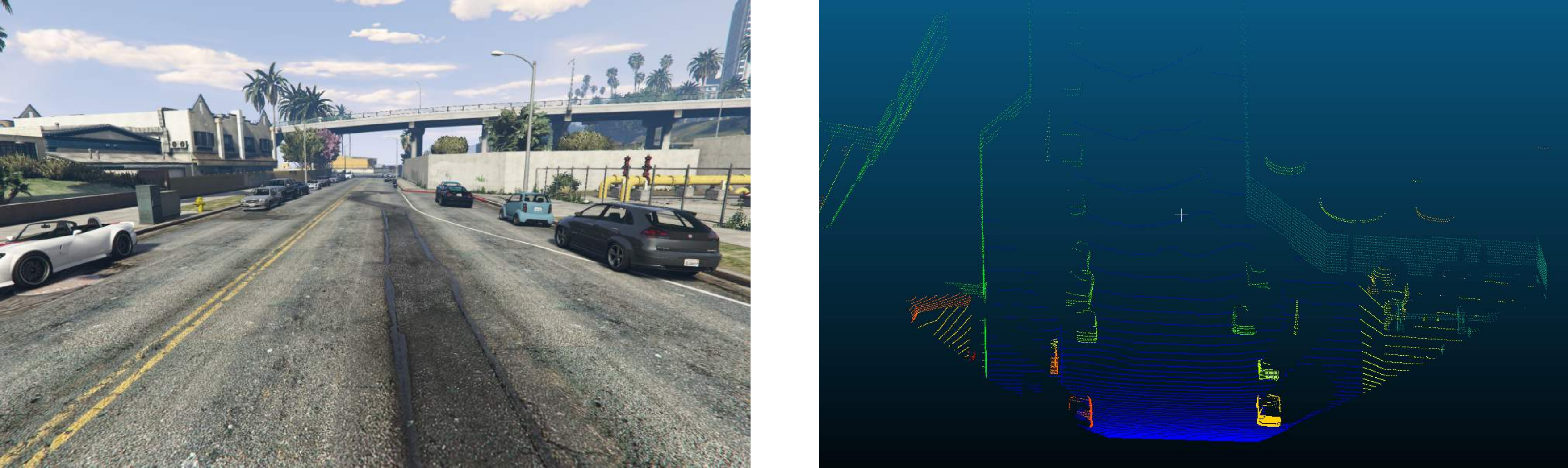}
    \caption{Left: Image of game scene from GTA-V. Right: LiDAR point cloud corresponding to the game scene.}
    \label{fig:GTA}
\end{figure}

\begin{figure}[h]
    \centering
    \includegraphics[width=3.3in,trim={0cm 4cm 0cm 3cm},clip]{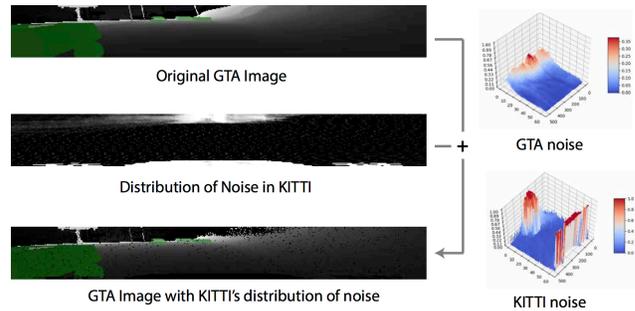}
    \caption{Fixing distribution of noise in synthesized data}
    \label{fig:noise}
\end{figure}

Using this simulator, we built a synthesized dataset with 8,585 samples, roughly doubling our training set size. To make the data more realistic, we further analyzed the distribution of noise across KITTI point clouds (Fig.~\ref{fig:noise}). We took empirical frequencies of noise at each radial coordinate and normalized to obtain a valid probability distribution: 1) Let $P_i$ be a 3D tensor in the format described earlier in Section \ref{sec:point_cloud} denoting the spherically projected ``pixel values'' of the $i$-th KITTI point cloud. For each of the $n$ KITTI point clouds, consider whether or not the pixel at the $(\tilde{\theta}, \tilde{\phi})$ coordinate contains ``noise.'' For simplicity, we consider ``noise'' to be missing data, where all pixel channels are zero. Then, the empirical frequency of noise at the $(\tilde{\theta}, \tilde{\phi})$ coordinate is
$$
\epsilon(\tilde{\theta}, \tilde{\phi}) = \frac{1}{n}\sum_{i=1}^n \mathds{1}_{\{P_i[\tilde{\theta}, \tilde{\phi}] = 0\}}.
$$
2) We can then augment the synthesized data using the distribution of noise in the KITTI data. For any point cloud in the synthetic dataset, at each $(\tilde{\theta}, \tilde{\phi})$ coordinate of the point cloud, we randomly add noise by setting all feature values to $0$ with probability $\epsilon(\tilde{\theta}, \tilde{\phi})$.

It is worth noting that GTA-V used very simple physical models for pedestrians, often reducing people to cylinders.  In addition, GTA-V does not encode a separate category for cyclists, instead marking people and vehicles separately on all accounts. For these reasons, we decided to focus on the ``car'' class for KITTI evaluation when training with our synthesized dataset.

\section{Experiments}
\subsection{Evaluation metrics}
We evaluate our model's performance on both class-level and instance-level segmentation tasks. For class-level segmentation, we compare predicted with ground-truth labels, point-wise, and evaluate precision, recall and IoU (intersection-over-union) scores, which are defined as follows:
\begin{gather*}
Pr_c = \frac{|\mathcal{P}_c \cap \mathcal{G}_c|}{|\mathcal{P}_c|}, 
recall_c = \frac{|\mathcal{P}_c \cap \mathcal{G}_c|}{|\mathcal{G}_c|}, 
IoU_c = \frac{|\mathcal{P}_c \cap \mathcal{G}_c|}{|\mathcal{P}_c \cup \mathcal{G}_c|},
\end{gather*}
where $\mathcal{P}_c$ and $\mathcal{G}_c$ respectively denote the predicted and ground-truth point sets that belong to class-$c$. $|\cdot|$ denotes the cardinality of a set. IoU score is used as the primary accuracy metric in our experiments.

For instance-level segmentation, we first match each predicted instance-$i$ with a ground truth instance. This index matching procedure can be denoted as $\mathcal{M}(i) = j$ where $i \in \{1, \cdots, N\}$ is the predicted instance index and $j \in \{\emptyset, 1, \cdots, M\}$ is the ground truth index. If no ground truth is matched to instance-$i$, then we set $\mathcal{M}(i)$ to $\emptyset$. The matching procedure $\mathcal{M}(\cdot)$ 1) sorts ground-truth instances by number of points and 2) for each ground-truth instance, finds the predicted instance with the largest IoU. The evaluation script will be released together with the source code.

For each class-$c$, we compute  instance-level precision, recall, and IoU scores as
\begin{gather*}
Pr_c = \frac{\sum_i|\mathcal{P}_{i,c} \cap \mathcal{G}_{\mathcal{M}(i),c}|}{|\mathcal{P}_c|}, \\
recall_c = \frac{\sum_i |\mathcal{P}_{i,c} \cap \mathcal{G}_{\mathcal{M}(i),c}|} 
{|\mathcal{G}_{c}|}, \\
IoU_c = \frac{\sum_i|\mathcal{P}_{i,c} \cap \mathcal{G}_{\mathcal{M}(i),c}|}{|\mathcal{P}_c \cup \mathcal{G}_c|}.
\end{gather*}
$\mathcal{P}_{i,c}$ denotes the $i$-th predicted instance that belongs to class-$c$. Different instance sets are mutually exclusive, thus $\sum_i|\mathcal{P}_{i,c}| = |\mathcal{P}_c|$. Likewise for $\mathcal{G}_{\mathcal{M}(i),c}$. If no ground truth instance is matched with prediction-$i$, then $\mathcal{G}_{\mathcal{M}(i),c}$ is an empty set. 

\subsection{Experimental Setup}
Our primary dataset is the converted KITTI dataset described above. We split the publicly available raw dataset into a training set with 8,057 frames and a validation set with 2,791 frames. Note that KITTI LiDAR scans can be temporally correlated if they are from the same sequence. In our split, we ensured that frames in the training set do not appear in validation sequences. Our training/validation split will be released as well. We developed our model in Tensorflow~\cite{TensorFlow} and used NVIDIA TITAN X GPUs for our experiments. Since the KITTI dataset only provides reliable 3D bounding boxes for front-view LiDAR scans, we limit our horizontal field of view to the forward-facing $90^\circ$. Details of our model training protocols will be released in our source code.

\subsection{Experimental Results}
Segmentation accuracy of SqueezeSeg is summarized in Table.\ref{tab:main}. We compared two variations of SqueezeSeg, one with the recurrent CRF layer and one without. Although our proposed metric is very challenging--as a high IoU requires point-wise correctness--SqueezeSeg still achieved high IoU scores, especially for the car category. Note that both class-level and instance-level recalls for the car category are higher than 90\%, which is desirable for autonomous driving, as false negatives are more likely to lead to accidents than false positives. We attribute lower performance on pedestrian and cyclist categories to two reasons: 1) there are many fewer instances of pedestrian and cyclist in the dataset. 2) Pedestrian and cyclist instances are much smaller in size and have much finer details, making it more difficult to segment. 

By combining our CNN with a CRF, we increased accuracy (IoU) for the car category significantly. The performance boost mainly comes from improvement in precision since CRF better filters mis-classified points on the borders. At the same time, we also noticed that the CRF resulted in slightly worse performance for pedestrian and cyclist segmentation tasks. This may be due to lack of CRF hyperparameter tuning for pedestrians and cyclists.

\begin{figure}[h]
    \centering
    \includegraphics[width=3.5in,trim={1cm 6cm 0cm 5.5cm},clip]{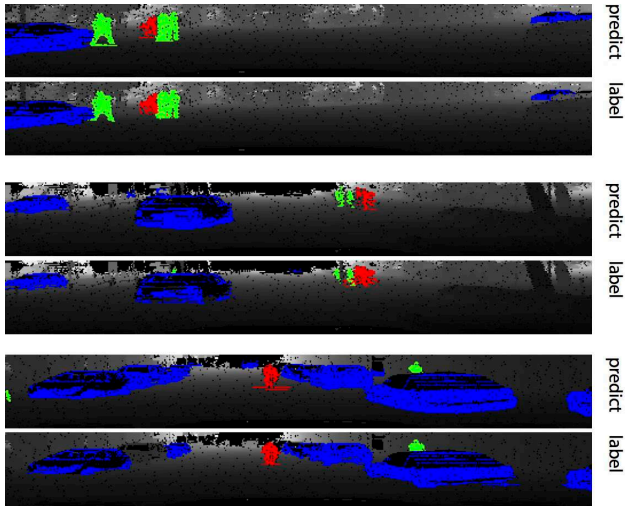}
    \caption{Visualization of SqueezeSeg's prediction on a projected LiDAR depth map. For comparison, visualization of the ground-truth labels are plotted below the predicted ones. Notice that SqueezeSeg additionally and accurately segments objects that are unlabeled in ground truth.}
    \label{fig:viz}
\end{figure}

\begin{table}[h]
\centering
\caption{Segmentation Performance of SqueezeSeg}
\label{tab:main}
\begin{tabular}{cc|ccc|ccc}
                                                 &         & \multicolumn{3}{c|}{Class-level} & \multicolumn{3}{c}{Instance-level} \\
                                                 &         & P         & R         & \textbf{IoU}      & P          & R         & \textbf{IoU}       \\ \hline
\multicolumn{1}{c|}{\multirow{2}{*}{car}}        & w/ CRF  & 66.7      & 95.4      & 64.6     & 63.4       & 90.7      & 59.5      \\
\multicolumn{1}{c|}{}                            & w/o CRF & 62.7      & 95.5      & 60.9     & 60.0       & 91.3      & 56.7      \\ \hline
\multicolumn{1}{c|}{\multirow{2}{*}{pedestrian}} & w/ CRF  & 45.2      & 29.7      & 21.8     & 43.5       & 28.6      & 20.8       \\
\multicolumn{1}{c|}{}                            & w/o CRF & 52.9      & 28.6      & 22.8     & 50.8       & 27.5      & 21.7      \\ \hline
\multicolumn{1}{c|}{\multirow{2}{*}{cyclist}}    & w/ CRF  & 35.7      & 45.8      & 25.1     & 30.4       & 39.0      & 20.6      \\
\multicolumn{1}{c|}{}                            & w/o CRF & 35.2      & 51.1      & 26.4     & 30.1       & 43.7      & 21.7      \\ \hline
\end{tabular}
\begin{tablenotes}
\small \item Summary of SqueezeSeg's segmentation performance. \textit{P, R, IoU} in the header row respectively represent precision, recall and intersection-over-union. IoU is used as the primary accuracy metric. All the values in this table are in percentages. 
\end{tablenotes}
\end{table}

Runtime of two SqueezeSeg models are summarized in Table.\ref{tab:runtime}. On a TITAN X GPU, SqueezeSeg without CRF only takes $8.7$ ms to process one LiDAR point cloud frame. Combined with a CRF layer, the model takes $13.5$ ms each frame. This is much faster than the sampling rate of most LiDAR scanners today. The maximum rotation rate for Velodyne HDL-64E LiDAR, for example, is 20Hz. On vehicle embedded processors, where computational resources are more limited, SqueezeSeg comfortably allows trade-offs between speed and other practical concerns such as energy efficiency or processor cost. Also, note that the standard deviation of runtime for both SqueezeSeg models is very small, which is crucial for the stability of the entire autonomous driving system. However, our instance-wise segmentation currently relies on conventional clustering algorithms such as DBSCAN\footnote{We used the implementation from 
\url{http://scikit-learn.org/0.15/modules/generated/sklearn.cluster.DBSCAN.html}}, which in comparison takes much longer and has much larger variance. A more efficient and stable clustering implementation is necessary, but it is out of the scope of this paper.

\begin{table}[h]
\centering
\caption{Runtime Performance of SqueezeSeg Pipeline}
\label{tab:runtime}
\begin{tabular}{c|cc}
\hline
                   & \begin{tabular}[c]{@{}c@{}}Average \\ runtime\\ (ms)\end{tabular} & \begin{tabular}[c]{@{}c@{}}Standard \\ deviation\\ (ms)\end{tabular} \\ \hline
SqueezeSeg w/o CRF & 8.7                                                               & 0.5                                                                  \\
SqueezeSeg         & 13.5                                                              & 0.8                                                                  \\
DBSCAN clustering            & 27.3                                                              & 45.8                                                                 \\ \hline
\end{tabular}
\end{table}

We tested our model's accuracy on KITTI data, when trained on GTA simulated data--the results of which are summarized in Table.\ref{tab:gta}. Our GTA simulator is currently still limited in its ability to provide realistic labels for pedestrians and cyclists, so we consider only segmentation performance for cars. Additionally, our simulated point cloud does not contain intensity measurements; we therefore excluded intensity as an input feature to the network. To quantify the effects of training on synthesized data, we trained a SqueezeSeg model on the KITTI training set, without using intensity measurements, and validated on the KITTI validation set. The model's performance is shown in the first row of the table. Compared with Table.\ref{tab:main}, the IoU score is worse, due to the loss of the intensity channel. If we train the model completely on GTA simulated data, we see significantly worse performance. However, combining the KITTI training set with our GTA-simulated dataset, we see significantly increased accuracy that is even better than Table.\ref{tab:main}. 

A visualization of the segmentation result by SqueezeSeg \textit{vs.} ground truth labels can be found in Fig.\ref{fig:viz}. For most of the objects, the predicted result is almost identical to the ground-truth, save for the ground beneath target objects. Also notice SqueezeSeg additionally and accurately segments objects that are unlabeled in ground truth. These objects may be obscured or too small, therefore placed in the ``Don't Care" category for the KITTI benchmark.

\begin{table}[h]
\centering
\caption{Segmentation Performance on the Car Category with Simulated Data}
\label{tab:gta}
\begin{tabular}{c|ccc|ccc}
\multicolumn{1}{l|}{} & \multicolumn{3}{c|}{Class-level}                                                  & \multicolumn{3}{c}{Instance-level}                                               \\
\multicolumn{1}{l|}{} & \multicolumn{1}{l}{P} & \multicolumn{1}{l}{R} & \multicolumn{1}{l|}{\textbf{IoU}} & \multicolumn{1}{l}{P} & \multicolumn{1}{l}{R} & \multicolumn{1}{l}{\textbf{IoU}} \\ \hline
KITTI                 & 58.9                  & 95.0                  & 57.1                              & 56.1                  & 90.5                  & 53.0                             \\
GTA                   & 30.4                  & 86.6                  & 29.0                              & 29.7                  & 84.6                  & 28.2                             \\
KITTI + GTA           & 69.6                  & 92.8                  & 66.0                              & 66.6                  & 88.8                  & 61.4                             \\ \hline
\end{tabular}
\end{table}

\vspace{-0.4cm}
\section{CONCLUSIONS}

We propose SqueezeSeg, an accurate, fast and stable end-to-end approach for road-object segmentation from LiDAR point clouds. Addressing the deficiencies of previous approaches that were discussed in the Introduction, our deep learning approach 1) does not rely on hand-crafted features, but utilizes convolutional filters learned through training; 2) uses a deep neural network and therefore has no reliance on iterative algorithms such as RANSAC, GP-INSAC, and agglomerative clustering; and 3) reduces the pipeline to a single stage, sidestepping the issue of propagated errors and allowing the model to fully leverage object context. The model accomplishes very high accuracy at faster-than-real-time inference speeds with small variance, as required for applications such as autonomous driving.
Additionally, we synthesize large quantities of simulated data, then demonstrate a significant boost in performance when training with synthesized data and validating on real-world data. We use select classes as a proof-of-concept, granting synthesized data a potential role in self-driving datasets of the future. 

\section*{Acknowledgement}
This work was partially supported by the DARPA PERFECT program, Award HR0011-12-2-0016, together with ASPIRE Lab sponsor Intel, as well as lab affiliates HP, Huawei, Nvidia, and SK Hynix. This work has also been partially sponsored by individual gifts from BMW, Intel, and the Samsung Global Research Organization.

\addtolength{\textheight}{-12cm}   





{\small
  \bibliography{bibliography.bib}
  \bibliographystyle{IEEEtran}
}

\end{document}